\documentclass{ecai} 

\usepackage{latexsym}
\usepackage{amssymb}
\usepackage{amsmath}
\usepackage{amsthm}
\usepackage{booktabs}
\usepackage{enumitem}
\usepackage{graphicx}
\usepackage{color}

\usepackage{verbatim}
\usepackage{float}
\usepackage{algorithm}
\usepackage{algorithmicx}
\usepackage[noend]{algpseudocode}

\usepackage{microtype}
\usepackage{graphicx}
\usepackage{subfigure}
\usepackage{booktabs}

\usepackage{url}
\usepackage{hyperref}

\newtheorem{theorem}{Theorem}

\newtheorem{corollary}[theorem]{Corollary}

\newtheorem{definition}{Definition}

\newcommand{\BibTeX}{B\kern-.05em{\sc i\kern-.025em b}\kern-.08em\TeX}

\begin{document}


\begin{frontmatter}


\paperid{123} 


\title{Neural Reward Machines}


\author[A]{\fnms{Elena}~\snm{Umili}\thanks{Corresponding Author. Email: umili@diag.uniroma1.it}}
\author[A]{\fnms{Francesco}~\snm{Argenziano}}
\author[A,B]{\fnms{Roberto}~\snm{Capobianco}} 

\address[A]{Sapienza University of Rome}
\address[B]{Sony AI}


\begin{abstract}
Non-markovian Reinforcement Learning (RL) tasks are very hard to solve, because agents must consider the entire history of state-action pairs to act rationally in the environment. Most works use symbolic formalisms (as Linear Temporal Logic or automata) to specify the temporally-extended task. These approaches only work in finite and discrete state environments or continuous problems for which a mapping between the raw state and a symbolic interpretation is known as a symbol grounding (SG) function. Here, we define Neural Reward Machines (NRM), an automata-based neurosymbolic framework that can be used for both reasoning and learning in non-symbolic non-markovian RL domains, which is based on the probabilistic relaxation of Moore Machines. We combine RL with semisupervised symbol grounding (SSSG) and we show that NRMs can exploit high-level symbolic knowledge in non-symbolic environments without any knowledge of the SG function, outperforming Deep RL methods which cannot incorporate prior knowledge. Moreover, we advance the research in SSSG, proposing an algorithm for analysing the groundability of temporal specifications, which is more efficient than baseline techniques of a factor $10^3$.
\end{abstract}

\end{frontmatter}


\section{Introduction}

Reinforcement learning (RL) tasks are traditionally modeled as Markovian Decision Processes (MDPs) \cite{sutton}, where task feedback depends solely on the last state and action. However, this formulation is inadequate for many decision problems, which are inherently non-Markovian or temporally extended \cite{non-markow-rewards1996, littman2017}. Intelligent agents tackling such tasks must consider the entire history of state-action pairs to act rationally within the environment. Current research in this field involves addressing non-Markovianity by expanding the state space with features that encode the environment history, and then solving the augmented-state problem with established RL algorithms.
The primary challenge lies in constructing these features. For non-symbolic-state problems, a popular approach combines RL algorithms with Recurrent Neural Networks (RNNs) \cite{RLRNN2, RLRNN3}, which automatically extract features from data sequences. While this method does not guarantee extracting Markovian features, it does not recquire any prior knowledge on the task.
In problems with a symbolic finite state space, most works utilize Linear Temporal Logic (LTL) \cite{LTL} or LTLf \cite{LTLf} to specify the temporal task. This specification is then compiled into an automaton, and the automaton state is combined with the environment state to render the decision process Markovian. Reward Machines (RMs) \cite{RM_journal} exemplify this approach. RMs can also be employed in non-symbolic-state environments if the symbol grounding (SG) function is known \cite{continuous_restr_bolt}. The latter maps the environment's raw state into a boolean interpretation over the symbols used by the specifications in LTL, making the symbols observable. In summary, RM-like methods presuppose prior knowledge of: (i) the SG function, and (ii) the temporal task specification in a logical formalism.
Despite previous work challenging the assumption of knowing the temporal specification \cite{reward_machine_learning_1, reward_machine_learning_2, reward_machine_learning_3, subgoalAutomaton}, no work thus far assumes a complete lack of knowledge regarding the SG function. We highlight that learning this function for realistic environments with raw and/or high-dimensional states can be challenging and necessitates a substantial amount of labeled data.

To achieve this, we propose a neurosymbolic (NeSy) approach to reward machines, which involves probabilistically relaxing logical constraints \cite{visual_reward_machines}. This approach integrates both logical reasoning and neural network-based perception into a unified framework, that we call "Neural Reward Machines" (NRM). We demonstrate the capability to relax the requirement of knowing the SG function while still leveraging available prior knowledge to enhance the performance of RL algorithms in non-symbolic non-Markovian environments. This achievement stems from combining RL on reward machines with semi-supervised symbol grounding (SSSG) \cite{DeepProbLog, LTN}.
SSSG consists in connecting some prior logical knowledge with raw data by supervising the formula output on the data. SSSG is often challenging, particularly when logical knowledge is vague or when connected data doesn't cover a diverse range of scenarios \cite{Umili2023KR, marconato2023RS}. In such cases, multiple solutions may exist — each maintaining consistency between knowledge and data - where, only one is the intended solution, while the others are deemed "Reasoning Shortcuts" (RS) \cite{marconato2023RS}.
In this paper, we also advance SSSG research by presenting an algorithm for identifying all RSs of a given temporal specification, assuming a complete dataset of observations. We call this kind of RSs \textit{unremovable}, since they solely depend on the structure of knowledge rather than the data used for SSSG or the specific symbol-perception task.

To the best of  our knowledge, this is the first work exploiting temporal logical knowledge in single RL tasks under the assumption of \textit{completely unknown} SG function, and proposing an algorithm for actually \textit{discovering} RSs of temporal specifications. 

\section{Related works}
\paragraph{Non-Markovian Reinforcement Learning with Temporal Specifications}
Temporal logic formalisms are widely used in Reinforcement Learning (RL) to specify non-Markovian tasks \cite{littman2017}. 
The large majority of works assumes the boolean symbols used in the formulation of the task are \textit{perfectly} observable in the environment \cite{reward-machine-sheila, restr_bolts, reward_machine_learning_1, reward_machine_learning_2, reward_machine_learning_3, subgoalAutomaton}. For this reason they are applicable only in symbolic-state environments or when a perfect mapping between the environment state and a symbolic interpretation is known, also called labeled MDP \cite{continuous_restr_bolt}.
Many works assume to know an \textit{imperfect} SG function for the task \cite{noisy_symbols_2020, noisy_symbols_2022, noisy_symbols_2022_shila}. Namely, a function that sometimes makes mistakes in predicting symbols from states or predicts a set of probabilistic \textit{beliefs} over the symbol set instead of a boolean interpretation. These works represent a step towards integration with non-symbolic domains. However, they do not address the problem of learning the SG function, but only how to manage to use a pre-trained imperfect symbol grounder.
Only one work in the literature assumes the same setting of ours \cite{ltl_no_grounding}, namely, that the agent observes sequences of non-symbolic states and rewards, and is aware of the LTL formula describing the task but without knowing the meaning of the symbols in the formula. This work employs a neural network composed of different modules, each representing a propositional symbol or a logical or temporal operator, which are interconnected in a tree-like structure, respecting the syntactic tree of the formula. This network learns a representation of states, that can be easily transferred to different LTL tasks in the same environment.
However, the key distinctions between our and their work are: (i) Kuo et al. \cite{ltl_no_grounding} learn a subsymbolic uninterpretable representation, while we learn \textit{precisely} a mapping between states and symbols; (ii) their method provides benefits only in a multitask setting and is unable to expedite learning on a single task, while ours learns and leverages the symbol grounding in the individual task.

\paragraph{Semisupervised Symbol Grounding}
Much prior work approaches the SG problem by assuming as inputs: (i) some prior symbolic logical knowledge, and (ii) a set of raw data that are annotated with the output of the knowledge (for example data are divided in positive and negative sample accordingly to the formula)\cite{ xu2018, LTN, DeepProbLog, Umili2023KR, abduction2019, huang_abduction_2021, Tsamoura_abduction_2021}. This practice is known as semi-supervised symbol-grounding (SSSG). SSSG is tackled mainly with two families of methods.
The first approach \cite{ xu2018, LTN, DeepProbLog, Umili2023KR} consists in embedding a continuous relaxation of the available knowledge in a larger neural networks system, 
and training the system so to align the outputs of the relaxation with the high-level labels.
The second approach \cite{abduction2019, huang_abduction_2021, Tsamoura_abduction_2021} instead maintains a crisp boolean representation of the logical knowledge and uses a process of logic \textit{abduction} to correct the current SG function, that is periodically retrained with the corrections.
Much work in this area does not take into account temporal aspects, and only few works focus on learning symbol grounding using temporal logics. In particular, \cite{Umili2023KR} employs an extension of Logic Tensor Networks (LTNs) to represent DFAs, which exploits for classifying sequences of images.
Our approach can be categorized in the first family of mentioned methods. Differently from \cite{Umili2023KR}, which employs LTNs and apply to classification domains, we use probabilistic Moore Machines and our main application is to RL. Furthermore our framework is more versatile than \cite{Umili2023KR}, as the temporal component can be both imposed and learned through the network.
\paragraph{Groundability and Reasoning Shortcuts}
In \cite{marconato2023RS}, the authors define reasoning shortcuts and outline a method for counting deterministic optima, assuming knowledge of: the logical knowledge, the support dataset, and the SG function. This method is inapplicable in our scenario since we assume the SG unknown. 
Conversely, \cite{Umili2023KR} introduces the concept of "ungroundability" 
as a property solely dependent on the logical knowledge and not on the data or the SG function. However, it gives only theoretical definitions and does not provide a method to actually discovering the RSs that cause the ungroundability. 
In this paper, we bridge these two concepts and we introduce the \textit{Unremovable Reasoning Shortcuts} (URS), as the RSs identifiable when considering all possible observations. Beyond offering a theoretical definition, we develop an algorithm for actually calculating URS for a temporal property, which is missing in the litterature. 
\section{Background}
\subsection{Moore Machines}
A Moore machine $M$ is a tuple $(P, Q, O,  q_0, \delta_t, \delta_o)$, where $P$ is a finite alphabet of input propositional symbols, $Q$ is a finite set of states, $O$ is a finite set of output symbols, $q_0 \in Q$ is the initial state, $\delta_t: Q \times P \to Q $ is the transition function, and $\delta_o: Q \to O $ is the output function.
Let $P^*$ be the set of all finite strings over $P$, and $\epsilon$ the empty string.
The transition function over strings $\delta_t^*:Q \times P^* \rightarrow Q$ is defined recursively as
\begin{equation}
\begin{array}{l} \label{eq:trans_over_strings}
    \delta_t^*(q,\epsilon) = q \\
    \delta_t^*(q, p+x) = \delta_t^*(\delta_t(q,p) , x)
\end{array}
\end{equation}
Where $p \in P$ is a symbol, $x \in P^*$ is a string, and $p+x$ is the concatenation of $p$ and $x$.
Consequently, the output function over strings $\delta_o^*:Q \times P^* \rightarrow O$ is defined as
\begin{equation} \label{eq:rew_over_strings}
    \delta_o^*(q, x) = \delta_o(\delta_t^*(q,x))
\end{equation}
We also define the string of output $\delta_o^{**}: Q \times P^* \rightarrow O^*$ as
\begin{equation}
    \delta_o^{**}(q, x) = [\delta_o^*(q, x|_1), \delta_o^*(q, x|_2), ..., \delta_o^*(q, x|_T)]
\end{equation}
Where $T$ is the length of string $x$, and $x|_i$ is the string $x$ up to index $i$.
Let $x_p=[p^{(1)}, p^{(2)},...,p^{(T)}]$ be the input string, where $p^{(t)}$ is the $t$-th character in the string, we denote as $x_q =[q^{(0)}, q^{(1)},...q^{(T)}]$ and as $x_o = [o^{(1)}, o^{(2)},...o^{(T)}]$ respectively the sequence of states and output symbols produced by the automaton while processing the string, namely $q^{(0)} =q_0$ and $q^{(t)}=\delta_t(q^{(t-1)},x^{(t)})$ and $o^{(t)}=\delta_o(q^{(t)})$  for all $t > 0$.
\subsection{Non-Markovian Reward Decision Processes and Reward Machines}
In Reinforcement Learning (RL) \cite{sutton} the agent-environment interaction is generally modeled as a Markov Decision Process (MDP).
An MDP is a tuple $(S,A,t,r,\gamma)$, where $S$ is the set of environment \textit{states}, $A$ is the set of agent's \textit{actions}, $t: S \times A \times S \rightarrow [0,1]$ is the \textit{transition function}, $r:S \times A \rightarrow \mathbb{R}$ is the \textit{reward function}, and $\gamma \in [0,1]$ is the \textit{discount factor} expressing the preference for immediate over future reward

In this classical setting, transitions and rewards are assumed to be Markovian – i.e., they are functions of the current state only.
Although this formulation is general enough to model most decision problems, it has been observed that many natural tasks are non-Markovian \cite{littman2017}.
A decision process can be non-markovian because markovianity does not hold on the reward function $r:(S\times A)^* \rightarrow \mathbb{R}$, or the transition function $t:(S \times A)^*\times S \rightarrow [0,1]$, or both. In this work we focus on Non-Markovian \textit{Reward} Decision Processes (NMRDP) \cite{restrainingBolts}.
Learning an optimal policy in such settings is hard, since the current environment outcome depends on the entire history of state-action pairs the agent has explored from the beginning of the episode; therefore, regular RL algorithms are not applicable. Rather than developing new RL algorithms to tackle NMRDP, the research has focused mainly on how to construct Markovian state representations of NMRDP. An approach of this kind are the so called Reward Machines (RMs).

RMs are an automata-based representation of non-Markovian reward functions \cite{RM_journal}. Given a finite set of propositions $P$ representing abstract properties or events observable in the environment, RMs specify temporally extended rewards over these propositions while exposing the compositional reward structure to the learning agent. Formally, in this work we assume the reward can be represented as a Reward Machine $RM = (P, Q, R,  q_0, \delta_t, \delta_r, L)$, where $P$ is the automaton alphabet, $Q$ is the set of automaton states, $R$ is a finite set of continuous reward values, $q_0$ is the initial state, $\delta_t: Q \times P \rightarrow Q$ is the transition function, $\delta_r: Q \rightarrow R$ is the reward function, and $L: S \rightarrow P$ is the labeling (or symbol grounding) function, which recognizes symbols in the environment states.
 
Let $x_s = [s^{(1)}, s^{(2)}, ... , s^{(t)}]$ be a sequence of states the agent has observed in the environment up to the current time instant $t$, we define the labeling function over sequences $L^*:S^* \rightarrow P^*$ as 
\begin{equation}
    L^*(x_s) = [L(s^{(1)}), L(s^{(2)}), ... , L(s^{(t)})]
\end{equation}
We denote with $\delta_t^*$ and $\delta_r^*$ the transition and reward function over strings, which are defined recursively, analogously to Equations \ref{eq:trans_over_strings} and \ref{eq:rew_over_strings}.
Given $x_s$, the RM produces an history-dependent reward value at time $t$, $r^{(t)} = \delta_r^*(q_0, L^*(x_s))$ and an automaton state at time $t$, $q^{(t)} = \delta_t^*(q_0, L^*(x_s))$. 
The reward value can be used to guide the agent toward the satisfaction of the task expressed by the automaton, while the automaton state can be used to construct a Markovian state representation. In fact it was proven that the augmented state $(s^{(t)}, q^{(t)})$ is a Markovian state for the task expressed by the RM \cite{restr_bolts}.

\paragraph{Example: Image-Based Minecraft-Like Environment} \label{par:example}
Consider the environment shown in Figure \ref{fig:minecraft}(a) consisting in a grid world containing different items: a pickaxe, a gem, a door and a lava-cell. The task to be accomplished in the environment is reaching at least once the pickaxe, the lava and the door cells, in any order. We can represent the task with the Moore Machine depicted in Figure \ref{fig:minecraft}(b), defined in terms of five symbols: one for each different item, plus a symbol indicating the absence of items. Therefore $P = \{$'pickaxe`, 'gem`, 'door`, 'lava`, 'empty-cell`$\}$. Each symbol is considered set to True when the agent is in the item-cell and false otherwise. The agent must learn how to navigate the grid in order to satisfy the task. At each step it receives as state $s \in S$ an image showing the agent in the grid, similar to that depicted in the figure. This state representation is not Markovian because from the current image the agent can recover only if it is located on a item \textit{now} and not which items it has visited in the past.


\section{Neural Reward Machines}
In this section, we formalize Neural Reward Machines, elaborate on their implementation using neural networks, and explore the reasoning and learning tasks achievable with NRMs. Subsequently, we will specifically delve into the integration of semi-supervised symbol grounding and RL through NRMs. Finally, we present our algorithm for identifying Unremovable Reasoning Shortcuts for a temporal property.
\subsection{Definition and Notations}
We define a Neural Reward Machine as a tuple $NRM = (S, P, Q, R, q_0, \delta_{tp}, \delta_{rp}, sg )$, where $S$ is the set of environment states, possibly infinite and continuous, the machine can process; $P$ is a finite set of symbols; $Q$ is a finite set of states, $R$ is a finite set of rewards; $q_0$ is the initial state; $\delta_{tp}: Q \times P \times Q \rightarrow [0,1]$ is the machine transition probability function; $\delta_{rp}: Q \times R \rightarrow [0,1]$ is the reward probability function and $sg: S \times P \rightarrow [0,1]$ is the \textit{symbol grounding} probability function.

\begin{figure*}[t]
    \centering
\subfigure[]{
    \includegraphics[width=0.24\textwidth]{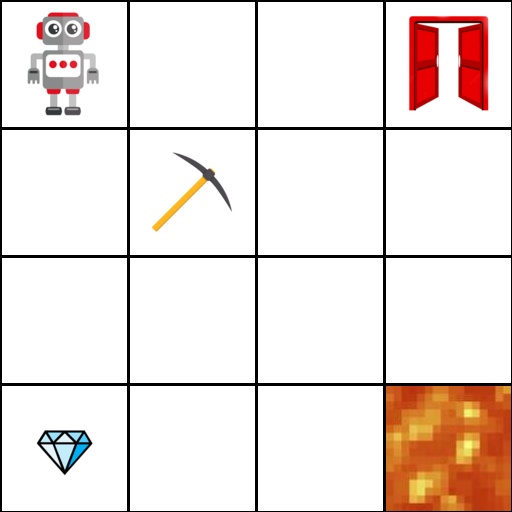}
    }
\subfigure[]{
    \includegraphics[width=0.3\textwidth]{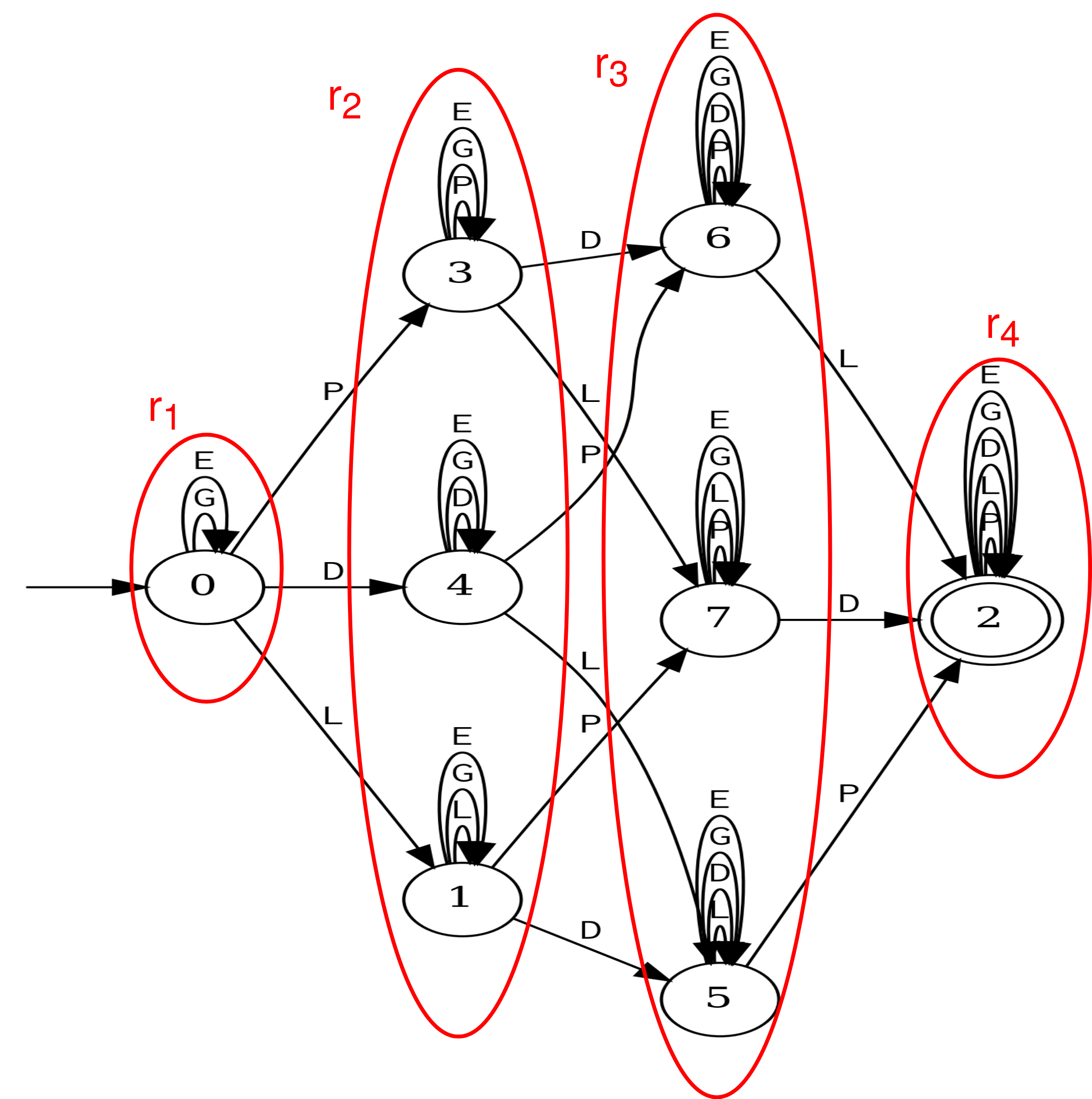}
    }
\subfigure[]{
    \includegraphics[width=0.41\textwidth]{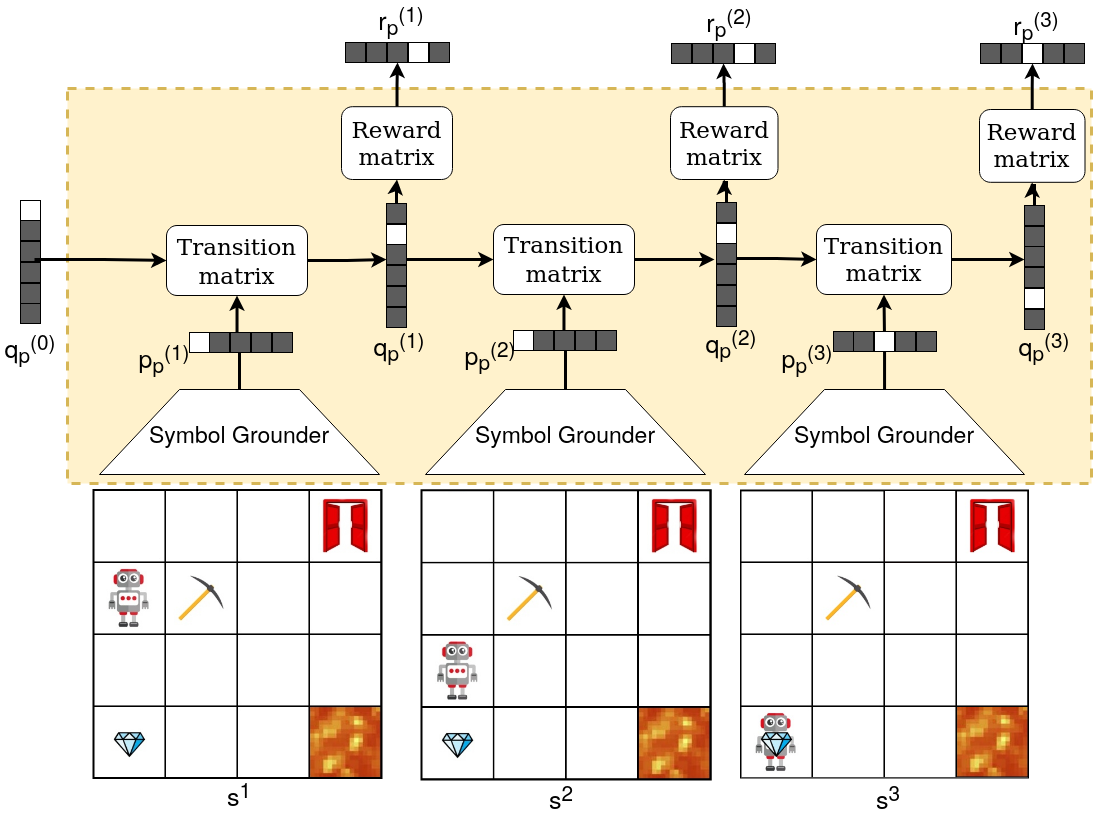}
}
    \caption{a) An example of non-Markovian navigation environment inspired by the Minecraft videogame. b) Moore Machine for the task: the agent has to visit the pickaxe (P), the lava (L) and the door (D) cells in any order. c) Implementation of NRM with neural networks.}
    \label{fig:minecraft}
\end{figure*}

Like an RM, an NRM produces a temporal state and a reward from a sequence of environment states. However, unlike an RM, the input sequences do not necessarily need to be symbolic; they can be of any type. The symbol grounding function maintains the link with the symbolic representation, assigning to a data instance $s \in S$ a probability value for each symbol in the alphabet $P$. 
Given a time sequence of environment observations $ x_s =[s^{(1)}, s^{(2)},...,s^{(T)}]$ the symbol grounding function grounds probabilistically each state in the set of symbols $P$ producing a sequence of symbol probability vectors $x_{p_p} = [p_p^{(1)}, p_p^{(2)},...,p_p^{(T)}]$, where the $k$-th component of $p_p^{(t)}$ is equal to $sg(s^{(t)}, p_k)$.
Note that probabilities are present both in the machine and the grounding: in our setting, the Reward Machine is a \textit{probabilistic} Moore Machine taking as input a \textit{probabilistic} symbols.
We represent the stochastic Moore machine in matrix representation, as:(i) the initial state probability vector $q_p^{(0)} \in [0,1]^{|Q|}$, containing at index $i$ a 1 if $q_0 = i$ and a 0 otherwise ; (ii) a transition matrix $M_t \in [0,1]^{|P|\times|Q| \times |Q|}$ containing at index $(p, q, q')$ the value of $\delta_{tp}(q, p, q')$; (iii) a reward matrix $M_r \in [0,1]^{|Q|\times|R|}$ representing the reward function and containing at index $(q,r)$ the value of $\delta_{rp}(q,r)$.
The sequence of probabilistic symbols returned by the grounder is processed by the probabilistic Moore Machine, which produces a sequence of state probability vectors $x_{q_p} = [q_p^{(1)}, q_p^{(2)},...q_p^{(T)}]$ and a sequence of reward probability vectors $x_{r_p} =[r_p^{(1)}, r_p^{(2)},...r_p^{(T)}]$.
\begin{equation} \label{eq:VRM_def}
\begin{array}{l}
    p_p^{(t)} = sg(s^{(t)}) \\
    q_p^{(t)} = \sum\limits_{i=1}^{i=|P|} p_p^{(t)}[i](q_p^{(t-1)}  M_t[i] ) \\ 
    r_p^{(t)} = q_p^{(t)} M_r \\ 
\end{array}
\end{equation}
Where we denote with $V[i]$ ($v[i]$), the component $i$ of matrix (vector) $V$ ($v$). We show the model in Figure \ref{fig:minecraft}(c).
Summarizing, given a time sequence of environment states $x_s$, the model in Equation \ref{eq:VRM_def} can be applied recursively on $x_s$ producing: a time sequence of input symbols $x_{p_p}$, one of machine states $x_{q_p}$, and a one of reward values $x_{r_p}$. Subscript $p$ denotes that sequences are \textit{probabilistically grounded}, namely, each point in the sequence is a probability vector.
While we denote with $x_p, x_q, x_r$, without subscript $p$, the \textit{symbolic} time sequences. Namely $x_p = [p^{(1)}, p^{(2)}, ..., p^{(T)}]$, $x_q = [q^{(1)}, q^{(2)}, ..., q^{(T)}]$, and, $x_r = [r^{(1)}, r^{(2)}, ..., r^{(T)}]$, with each point in the sequence being a symbol in the finite set $P$, $Q$ and $R$ respectively. Symbolic sequences can also be obtained from the probabilistic ones choosing the most probable symbol at each time step. Namely $p^{(t)} = p_i\in P$, with $i = \arg\max_i(p_p^{(t)}[i])$; $q^{(t)} = q_j\in Q$, with $j = \arg\max_j(q_p^{(t)}[j])$; $r^{(t)} = r_k \in R$, with $k = \arg\max_k(r_p^{(t)}[k])$.
We next build our neural framework by using continuous parametric models based on this structure.

\subsection{Implementation}
In this section, we discuss the implementation of the formalism described above using Neural Networks (NN). Notably, we employ probabilistic relaxations and matrix representations of functions to facilitate this integration.
Neural Networks are inherently continuous as they learn through a gradient-based optimization process, therefore they can face challenges in learning a `crispy' boolean logical model like a Moore Machine. In our case, we aim for each function in the NRM definition—namely, $sg$, $\delta_{tp}$, and $\delta_{rp}$—to be both \textit{initialized} with prior knowledge and \textit{learned} from data.

For the SG function, it can take any form, with the only constraint being that it must return a probability vector. Therefore, it can be implemented with any NN featuring softmax activation on the output layer.
Conversely, learning the Moore Machine with backpropagation is more challenging. Our approach is based on the intuition that Probabilistic Moore Machines (PMM) are closely related to Recurrent Neural Networks (RNN), as they calculate the next state and output using multiplications between continuous vectors and matrices, similar to RNNs. Even though (deterministic) machines can be represented in matrix form with only one-hot row vectors.
Following this idea, we define the recurrent module as a parametric PMM whose representation can be driven to be close to one-hot during training, enabling logical induction through backpropagation. We achieve this effect using an activation function that smoothly approximates discrete output, as seen in prior work \cite{gumbel_softmax, walke_ltl, marine_rule_induction}.
In particular, we use a modified version of the classical softmax activation function: $softmax_\tau(x, \tau) = softmax(x/\tau)$, where $0 < \tau \leq 1$ is a temperature value controlling the activation steepness. When $\tau$ is close to 0, the $\tau$-softmax returns outputs close to one-hot vectors; when $\tau$ is 1, the function is the standard softmax.

The NRM implementation with parametric models is represented as follows:
\begin{equation}
\begin{array}{l} \label{eq:VRM_net}
p_p^{(t)} = sg(s^{(t)}; \theta_{sg})  \\
q_p^{(t)} = \sum\limits_{i=1}^{i=|P|} p_p^{(t)}[i](q_p^{(t-1)} (softmax_\tau(M_t , \tau))[i]) \\
r_p^{(t)} = q_p^{(t)}softmax_\tau(M_r , \tau) \\

\end{array}
\end{equation}
where the symbol grounding function $sg$ is now a parametric function with learnable parameters $\theta_{sg}$, and $M_t$ and $M_r$ are matrices of parameters. We process $M_t$ and $M_r$ with $softmax_\tau$ before using them in order to ensure they tend to represent a (deterministic) Moore Machine as the temperature decreases. The discretization of parameters in $M_t$ and $M_r$ through temperature decreasing is unnecessary and can be omitted if the machine is initialized with a known Moore Machine. It is only required when \textit{learning} the machine from data. A scheme of the proposed implementation is shown in Figure \ref{fig:minecraft}(c).
The model in equation \ref{eq:VRM_net} can be seen as three parametric functions (NNs) producing the three sequences of probability vectors as output, when given a sequence of states $x_s$ as inputs, namely: 
$x_{p_p} = \mathcal{G}(x_s | \theta_{sg})$, $x_{q_p} = \mathcal{Q}(x_s | \theta_{sg}, M_t)$, and $x_{r_p} = \mathcal{R}(x_s | \theta_{sg}, M_t, M_r)$.
\subsection{Reasoning and Learning with NRM}
As anticipated the framework is designed to allow both reasoning and learning and both initialization with background knowledge and training with data of each module of the machine. Specifically, different configurations of prior knowledge on the NRM parameters ($\theta_{sg}$, $M_t$, and $M_r$) and of the possibility to observe rewards in the environment or not result in different possible uses of NRMs, that we describe here.

\paragraph{Pure Reasoning}
By feeding the machine with a time sequence of states $x_s$ observed in the environment, we can calculate the sequence of probabilities on the machine states, $x_{q_p} = \mathcal{Q}(x_s | \theta_{sg}, M_t)$, and reward values, $x_{r_p} = \mathcal{R}(x_s | \theta_{sg}, M_t, M_r)$ which can be used in the same manner as RMs do: for designing non-Markovian reward functions, for augmenting the environment state, for Counterfactual Experience or hierarchical RL \cite{RM_journal}. This process requires prior knowledge of \textit{all} the NRM parameters $\theta_{sg}$, $M_t$, and $M_r$ and does not require any reward feedback from the environment.

\paragraph{Pure Learning}
In case we do not know any of the NRM parameters but we can observe in the environment sequences of states $x_s$ and ground truth sequences of rewards $x_r$ we can exploit the latter to learn the NRM modules entirely from data. It is obtained by minimizing the cross-entropy loss (or other classification loss functions) between the rewards probabilities predicted by the NRM and the ground-truth rewards.
\begin{equation} \label{eq:loss_function}
    L(x_s, x_r) = \text{cross-entropy}(\mathcal{R}(x_s | \theta_{sg}, M_t, M_r), x_r)
\end{equation}

\paragraph{Learning and Reasoning Integration}
If only \textit{some} of the NRM modules are known, and sequences of states and rewards are observable in the environment, we can train the missing modules to align with both prior knowledge and data. This entails initializing the known modules with the available knowledge and training the remaining parameters using loss function in Equation \ref{eq:loss_function}. Various configurations are feasible in this scenario. Here, we specifically focus on \textbf{symbol grounding}: learning $\theta_{sg}$ while assuming $M_t$ and $M_r$ are known. Another reasoning-learning configuration, which we are currently exploring, is that of \textbf{automata learning}, where $\theta_{sg}$ is known while $M_t$ and $M_r$ are unknown, which we let for future research.

\subsection{Exploiting NRMs for non-Markovian RL} \label{sec:nrm_and_RL}

In this paper, we assume that an agent has to learn to perform a temporally extended task by optimizing a non-Markovian reward signal. As is standard in RL, the agent interacts with the environment by taking an action and it receives a new state of the environment and a new reward in response. Additionally, we assume that the agent has a \textit{purely symbolic} knowledge of the task it is executing. For example, in a scenario like the Minecraft game depicted in Figure \ref{fig:minecraft}(a), the task for the agent could be "reach the door, the pickaxe, and the lava in any possible order." In this case, the agent would have access to the Moore Machine shown in Figure \ref{fig:minecraft}(b). However, it lacks knowledge on \text{how to recognize} a pickaxe, a gem, or a door within its environment. In this setting, traditional RMs are inapplicable because of the missing symbol grounding function, while purely deep-learning-based approaches have no way to exploit the symbolic knowledge and will rely only on rewards.
With our framework we exploit both symbolic knowledge and data from the environment by interleaving RL and semi-supervised symbol grounding of the NRM.
In particular, we initialize the parameters $M_t$ and $M_r$ with the task prior knowledge, and the parameters of the grounder $\theta_{sg}$ with random weights. We record each episode of interaction, as a sequence of states $x_s$ and a sequence of rewards $x_r$. We calculate the machine state sequence $x_{q_p}$ with the NRM as $\mathcal{Q}(x_s| \theta_{sg}, M_t, M_r)$, and we augment each environment state with the predicted machine state (\textbf{reasoning step}). The interaction is leaded by a RL algorithm running on the augmented state $(s^{(t)}, q_p^{(t)})$.
Since the symbol grounder is initially randomized, the predicted machine state initially deviates from the ground truth, and the state representation is not perfect. However, at regular intervals, we update the symbol grounder by training on sequences $x_t$ and $x_r$ collected in the environment so to minimize the loss in Equation \ref{eq:loss_function} (\textbf{learning step}).
As the agent observes more scenarios in the environment, the symbol grounding function of the model gradually becomes more similar to the unknown ground truth, and so does the distribution of machine states. In case the perfect SG function is learned the NRM becomes equivalent to a RM.

\paragraph{Reward Values as Supervision}
Since the reward values serve as supervision for the NRM, we have observed that overly sparse rewards are not very effective for symbol grounding. For instance, consider the automaton shown in Figure \ref{fig:minecraft}(b), with 8 states, of which only one is final (state 2). If we employ a sparse signal, rewarding the agent solely upon task completion, this entails assigning a positive reward to state 2 and 0 reward to all the other states. Subsequently, all the unsuccessful episodes would provide no feedback to the symbol grounder. Moreover, in case the agent accomplishes the task, the grounder would glean feedback solely to discern that the \textit{last} observation leading to winning reward ought to be associated with one of the three symbols bringing from a non rewarding state to state 2: lava (L), door (D), or pickaxe (P). However, it would lack the means to distinguish among these three symbols in the last observation, or discern which symbols the agent has encountered in the previous observations.
We, therefore, assume a potential-based reward shaping \cite{reward_shaping_potential}, where the potential function varies based on the distance on the automaton from the nearest final state. This partitions the states of the machine in figure according to 4 reward values, $r_1, r_2, r_3$ and $r_4$. Reward shaping is very commonly applied in RM applications to facilitate policy learning through RL \cite{RM_journal}. Note that this type of reward generally does not provide distinct feedback for each single state of the machine. Consequently, while the resulting grounding process is feasible, it is by no means straightforward.

\subsection{Groundability Analysis of Temporal Specifications}
 In our approch RL and Symbol Grounding are closely intertwined and mutually influence each other. SG relies on RL as we collect data through RL exploration for SG training. Simultaneously, RL relies on SG, as it utilizes an estimated representation of the automaton's state, which becomes more accurate as SG accurately predicts symbols.
 At the same time, regardless of the specific application, all types of semisupervised SG are affected by some reasoning shortcuts, especially when the knowledge is somewhat `trivial' - including formulas that are either trivially false or trivially true across most instances - or contains specific symmetries \cite{Umili2023KR}. 

Therefore, in this section, we delve into the issue of \textit{ungroundability}. We first review the definitions of ungroundability and reasoning shortcuts and we then  device our own algorithm to find all the RS of a given temporal specification depending solely on the specifics, that we call for this reason 'unremovable`.
\paragraph{Reasoning Shortcuts}
Given the problem of grounding symbols of the alphabet $P$ of a certain logical knowledge $\phi$ exploiting the knowledge and a dataset $D$ of couples (data, output of the formula), a RS is a renaming of symbols, $\alpha:P \rightarrow P$, different from the identity, $ \exists p \in P \vert \alpha(p) \neq p$, which maintains perfect alignment of $\phi \circ \alpha$ with the data in $D$. In \cite{marconato2023RS} they estimate the number of reasoning shortcuts by assuming prior knowledge of the groundtruth SG function $sg^*$, the logical knowledge $\phi$ and the support dataset $D$.
\begin{equation}
\#RS(\phi, D, sg^*) = \sum_{\alpha \in \mathcal{A}} \mathbf{1} \left\{ \bigwedge_{d \in D} (\phi \circ \alpha)(sg^*(d)) = \phi(sg^*(d)) \right\}
\end{equation}

Where $\mathcal{A}$ is the set of all possible mappings $\alpha$.
Only the $\alpha$ mapping each symbol to itself is indeed the correct solution, therefore the knowledge is said to admit RSs on D if $\#RS(\phi, D, sg^*) > 1$

\paragraph{Ungroundability}
The following definition of ungroundability is given in \cite{Umili2023KR}: a symbol \( p_i \in P\) is ungroundable through a formula \( \phi \) if there exists a mapping \( \alpha: P \rightarrow P \) that maps \( p_i \) to \( p_j\) with \( p_i \neq p_j \) such that \( \phi \equiv (\phi \circ \alpha) \).
Where \( \equiv \) denotes logical equivalence.
Notice that ungroundability is a property of \textit{a symbol} in the \textit{formula}, and does not depend on the particular supporting dataset used in the application, nor it needs the knowledge of $sg^*$ to be verified.
However, \cite{Umili2023KR} does not define a method to actually \textit{find} the mappings $\alpha$ which maintain logical equivalence between $\phi$ and $\phi \circ \alpha$ for a specific $\phi$. Nonetheless, these mappings could be identified using a naive brute-force method. This involves examining the semantic equivalence between $\phi$ and $\phi \circ \alpha$ for every possible mapping $\alpha \in \mathcal{A}$ and returning the mapping that yields a positive check. Given that the number of possible mappings is $|\mathcal{A}|=|P|^{|P|}$ and equivalence checking has an exponential cost, the brute-force approach is very time consuming and can become impracticable for long formulas defined over large alphabets.
Here, we introduce a smarter algorithm for identifying knowledge-preserving mappings. This algorithm leverages certain stopping criteria, which we will delineate later, to enhance speed by a factor of $10^3$. We term the mappings discovered by this algorithm as \textit{Unremovable Reasoning Shortcuts} (URS), referring to the RS that are \textit{pathological} of the logical knowledge, which would be never eliminated, even considering an ideal support dataset including all the possible observations of the domain.

\paragraph{Unremovable Reasoning Shortcuts (URS)}
To calculate URSs we focus on the complete support $D^*$. Notice that, since the support is complete, we can directly create synthetically the complete \textit{symbolic} dataset $D_{sym}^*$, and count the number of RS on that without knowing the ground truth symbol grounding function $sg^*$.
We define the number of reasoning shortcuts under complete support assumption as 
\begin{equation} \label{eq:URS}
    \#RS^*(\phi) =  \sum_{\alpha \in \mathcal{A}} \mathbf{1} \left\{ \bigwedge_{d_{sym} \in D_{sym}^*} (\phi \circ \alpha)(d_{sym}) = \phi(d_{sym}) \right\}
\end{equation}
\paragraph{Finding URS of a Temporal Property}
From now on, we assume that the logical knowlegde $\phi$ is a Moore Machine, or equivalently a Deterministic Finite Automaton (DFA) (DFAs indeed can be considered MMs with a binary output alphabet), an LTLf, or LDLf formula (which can be automatically translated into DFAs \cite{LTLf}).
It is worth to notice that \textit{temporal} formalisms (as LTLf, MMs, etc) do not admit a \textit{finite} complete symbolic support $D_{sym}^*$, because strings in any support can always be extended at will in the time dimension.
Therefore we cannot verify Equation \ref{eq:URS} directly.
We present in Algorithm \ref{alg:find_RS} our algorithm for finding URS, which is basically an efficient way to verify Equation \ref{eq:URS} on a \textit{finite} symbolic support that can still be considered \textit{complete} for $\phi$. We find this kind of support applying the following theorems.

\begin{theorem}\label{th:1}
Let $D^*_{sym}(L)$ denote the set of all strings over $P$ with maximum length equal to $L$. If $L_1 \leq L_2$, then $\#RS^*(\phi, D^*_{sym}(L_2)) \leq \#RS^*(\phi, D^*_{sym}(L_1))$.  
\end{theorem}
\begin{corollary} \label{ch:1}
The number of RS calculated on any complete support with finite horizon $L$ is an upper bound for the number of URS.
$ \#RS^*(\phi) \leq \#RS^*(\phi, D^*_{sym}(L))$ $\forall L < \infty$.
\end{corollary}
\begin{theorem} \label{th:2}
Let $A \subseteq Q$ be the set of absorbing states of $\phi$. \footnote{A state $q \in Q$ is called absorbing if $\delta(q,p) = q$ $\forall p \in P$}. If: (i) a string $x$ has reached an absorbing state, $\delta_t^*(q_0, x) \in A$. (ii) for a certain mapping $\alpha$, also $x \circ \alpha$ has reached an absorbing state, $\delta_t^*(q_0, x \circ \alpha) \in A$. (iii) $\alpha$ is a working map for $x$, namely $\delta_o^{**}(q_0, x) = \delta_o^{**}(q_0, x \circ \alpha)$.
Then $\alpha$ is a working map also for all the strings having $x$ as prefix, namely $\delta^{**}_o(q_0, x + y) = \delta^{**}_o(q_0, (x + y) \circ \alpha) \forall y \in P^*$.
\end{theorem}
\begin{theorem} \label{th:3}
Given that: (i) $\alpha$ is a working map, for the string $x+z$, with $x,z \in P^*$, $\delta_o^{**}(q_0, x+z) = \delta_o^{**}(q_0, (x+z) \circ \alpha)$; (ii) $\exists p \in P$ such that both $x$ and $x+p$ reach the same state and $x \circ \alpha$ and $(x+p) \circ \alpha$ reach the same state, $\delta_t^*(q_0, x) = \delta_t^*(q_0, x+p) = q \quad \wedge \quad \delta_t^*(q_0, x) = \delta_t^*(q_0, x+p)$.
Then $\alpha$ works for the string $x+y+z$, $\delta_o^{**}(q_0, x+y+z) = \delta_o^{**}(q_0, (x+y+z) \circ \alpha)$, $\forall y \in p^*$ \footnote{with $p^*$ being the set of strings obtained repeating the symbol $p$ an arbitrary number of times.}.
\end{theorem} 
We report the proofs of Theorems \ref{th:1}, \ref{ch:1}, \ref{th:2}, and \ref{th:3} in the appendix.
Algorithm \ref{alg:find_RS} keeps a dataset D[$\alpha$] for each candidate URS $\alpha$. The set of candidates URS is initialized with $\mathcal{A}$. D[$\alpha$] is initialized with all the possible strings on $P$ with length one for each $\alpha$. At each iteration we verify if $\alpha$ is a working for dataset D[$\alpha$] (namely if Equation \ref{eq:URS} with D[$\alpha$] in place of $D_{sym}^*$ and $\alpha$ in place of $\mathcal{A}$ gives 1). If it is working it remains a candidate otherwise is discarded. Then we try to extend the datasets of surviving maps of one step. This follows Theorem 1, which says that if a map does not work for a shorter complete dataset has no chance to work in the longer one. Following theorems 3 and 4, we extend a string $x$ in D[$\alpha$] with the symbol $p \in P$ only if: (i) not both $x$ and $x \circ \alpha$ bring to an absorbing state (Theroem 3) and (ii) not both state(x) = state(x+p) and state($x \circ \alpha$) = state($(x+p) \circ \alpha$) (theorem 4).
We iterate until at least one dataset has grown with respect to the past iteration. Finally we return the $\alpha$ remained in the set of candidates as URSs.

\section{Experiments}
\begin{algorithm}[tb]
\caption{Find\_unremovable\_reasoning\_shortcuts($\phi$)}
\begin{algorithmic}
\State URScandidates $\gets$ $\mathcal{A}$
\For{$\alpha$ in URScandidates}
    \State D[$\alpha$] $\gets$ $\phi$.P
\EndFor
\While{URScandidate not empty}
    \For{$\alpha$ in URScandidate}
        \State D\_next\_$\alpha$ $\gets$ \{\}
        \If{$\alpha$ is working on D[$\alpha$]}
            \For{x in D[$\alpha$]}
                \If{not $\phi$(x).q.abs or not $\phi$($\alpha$(x)).q.abs}
                    \For{p in $\phi$.P}
                        \State $x' \gets x + p$
                        \If{ $\phi$(x).q $\neq$ $\phi$(x').q or $\phi$($\alpha$(x)).q $\neq$ $\phi$($\alpha$(x')).q}
                            \State D\_next\_$\alpha$ $\gets$ D\_next\_$\alpha$ + x'
                        \EndIf
                    \EndFor
                \EndIf
            \EndFor
        \EndIf
        \If{D\_next\_$\alpha$ is empty}
            \State URScandidate $\gets$ URScandidate - $\alpha$
        \Else
            \State D[$\alpha$] $\gets$ D\_next\_$\alpha$
        \EndIf
    \EndFor
\EndWhile
\State \Return URScandidates
\end{algorithmic}
\label{alg:find_RS}
\end{algorithm}

\begin{figure*}
        \centering

\subfigure[]{
    \includegraphics[width=0.235\textwidth]{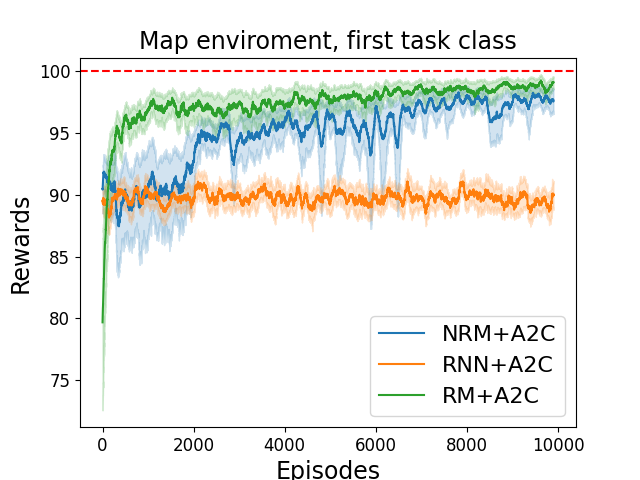}
    }
\subfigure[]{
    \includegraphics[width=0.235\textwidth]{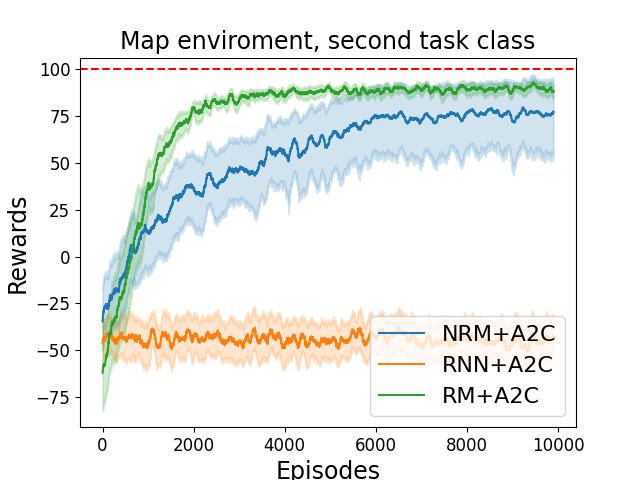}
    }
\subfigure[]{
    \includegraphics[width=0.235\textwidth]{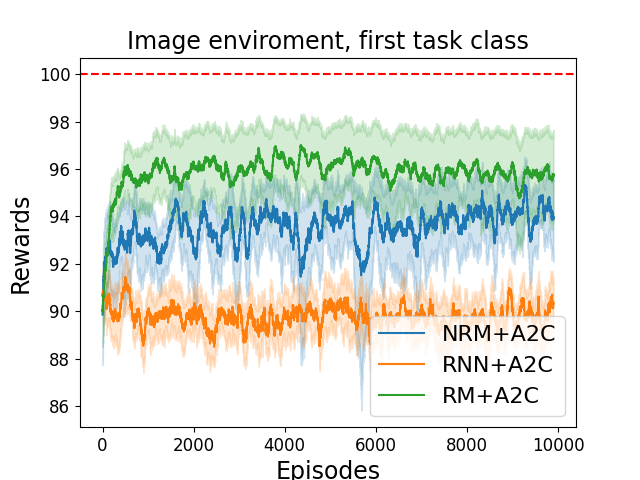}
}
\subfigure[]{
    \includegraphics[width=0.235\textwidth]{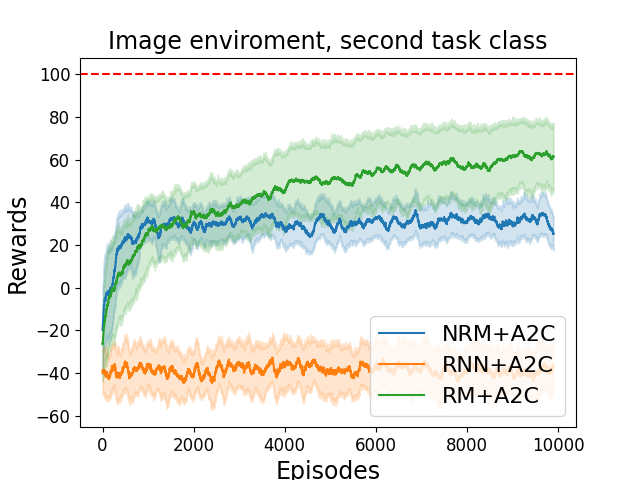}
}

    \caption{Results in the map environment when using a) tasks of the first class, and b) in the second class. Results in the image environment when training on c) the first task class d) the second task class.}
    \label{fig:results}
\end{figure*}
In this section, we report the experiments validating our framework. Our code is available on github: \href{https://github.com/KRLGroup/NeuralRewardMachines}{https://github.com/KRLGroup/NeuralRewardMachines}.

\paragraph{Application}
We test NRMs on two types of environment inspired by the Minecraft videogame, similar to those considered by \cite{stochastic_reward_machines} and \cite{noisy_symbols_2022_shila} and decribed in the example of Section \ref{par:example}, in which we assume the SG function is unknown.

\paragraph{States}
Based on this application we construct two environments that showcase different levels of difficulty in terms of symbol grounding:
(i) \textbf{The map environment}, in which the state is the 2D vector containing the x and y current agent location, (ii) \textbf{The image environment}, in which the state is an image of 64x64x3 pixels showing the agent in the grid, like the one shown in Figure \ref{fig:minecraft}(a).

\paragraph{Rewards}
We express the non-Markovian reward as an LTLf formula, that is then transformed into a DFA and, ultimately, into a Moore Machine. In the last process, each state $q$ is assigned a reward value, which is maximum if $q$ is a final states and gradually decreases as the distance from $q$ to the final state increases (as detailed in Section \ref{sec:nrm_and_RL} and shown in Figure \ref{fig:minecraft}(b)). Reward values are scaled so to give maximum cumulative reward always equal to 100.
We focus on patterns of formulas that are popular in non-Markovian RL \cite{RM_journal, LTL2action}, that we call here using the categorization given in \cite{ltl_patterns}.
\textbf{Visit} formulas: we denote as Visit($p_1$, $p_2$, ... , $p_n$) the LTLf formula F $p_1$ $\wedge$ F $p_2$ $\wedge$ ... $\wedge$ F $p_n$, expressing that the agent has to make true at least once each $p_i$ in the formula in any possible order.
\textbf{Sequenced Visit}: we denote with Seq\_Visit($p_1$, $p_2$, ... , $p_n$) the LTLf formula F($p_1 \land$ F($p_2 \land \ldots \land$ F($p_n$))), expressing that we want the symbol $p_{i+1}$ made true at least once \textit{after} the symbol $p_i$ has become True. \textbf{Global Avoidance}: we denote with Glob\_Avoid($p_1$, $p_2$, ... , $p_n$) the formula G ($\neg p_1$) $\wedge$ G ($ \neg p_2$) $\wedge$ ... $\wedge$ G ($ \neg p_n$), expressing that the agent must always avoid to make True the symbols $p_1$, $p_2$, ... , $p_n$. F and G are temporal operators called respectively `eventually' and `globally', we refer the reader to \cite{LTL} for the formal semantics of these operators that we explain here only intuitively. Based on these patterns of formulas we construct tasks of increasing difficulty that we aggregate in two classes. The \textbf{first class} is a set of tasks obtained as a conjunction of visit and sequenced visit formulas. The \textbf{second class} contains conjunctions of visit, sequenced visit and global avoidance formulas. We report in the appendix the formulas considered.

\paragraph{Comparisons} 
Given the absence of alternative methods in the literature for leveraging ungrounded prior knowledge in non-Markovian RL, we compare NRM with the two most popular baselines in this domain: RNNs and RMs. However, it is important to note that the three methods assume different levels of knowledge.
Specifically, RM's performance serves as an upper bound for NRMs, as in the optimal scenario, where NRMs precisely learn the ground-truth SG function (which RMs possess from the outset), 
they become equivalent to RMs.
We use Advantage Actor-Critic (A2C) \cite{mnih2016asynchronous} as RL algorithm for all the methods. Hence we denote the methods (and which information they can exploit for learning): \textbf{RNN+A2C} (rewards) the baseline based on RNNs,  \textbf{NRM+A2C} (rewards+symbolic task) our method, and \textbf{RM+A2C} (rewards+symbolic task+grounding) the upper bound obtained with RMs.

\begin{table}[t]
    \centering
    \begin{tabular}{lcccc}
        \toprule
        Task & \#RS* & Alg. 1 exec. time (s) & Brute force  exec. time (s)\\
        \midrule
        Class 1 & $27.3 \pm 17.8$ & $0.6 \pm 0.4$ & $434.6 \pm 195.6$\\
        Class 2 & $6 \pm 2$ & $0.2 \pm 0.0$ & $399.7 \pm 24.5$ \\
        \bottomrule
    \end{tabular}
    \caption{Groundability analysis of formulas selected as RL tasks}
    \label{tab:groundability}
\end{table}
\paragraph{RL Results}
In Figure \ref{fig:results}, we show the training rewards obtained by the three methods on both the map environment for the first task class (a) and the second task class (b), and the image environment for the first (c) and second (d) task classes. For each task and method, we perform 5 runs with different random seeds.
The figure clearly shows that the performance of NRMs falls between that of RNNs and RMs, closely resembling the latter across various experimental configurations, despite assuming less prior knowledge than RMs. This demonstrates that NRMs can effectively leverage the incomplete prior knowledge they possess. Specifically, NRMs converge almost to the same rewards as RMs, especially in the map environment. Generally, RNNs perform much worse than the other methods, often quickly converging to a local maximum of the reward, especially in challenging temporal tasks (class 2).
All methods share the same hyperparameter settings for A2C and the neural networks used for the policy, the value, and the feature extraction (only used in the image environment), which we detail in the appendix along with information on the network used as the grounder in the two environments.

\paragraph{Groundability Analysis}

Finally, we report the number of URS identified running the Algorithm \ref{alg:find_RS} on the selected tasks, along with the execution time of our algorithm and that of a brute-force version that simply iterates over each possible map $\alpha$ and checks whether $\phi \equiv (\phi \circ \alpha)$. Results are reported in Table \ref{tab:groundability}.
The number of URS found confirms our intuition that, as the tasks' difficulty increases, the number of URS decreases, resulting in a reduction in groundability difficulty. This difficulty had never been investigated before and it is hard to estimate. We start exploring it using the URS tool, showing that the two task classes have significantly different average $\#RS^*$.
However, note that the number of URSs for the chosen tasks is always relatively low (27 out of a maximum of $|P|^{|P|}$ = 3125 for 5 symbols), which indicates the tasks chosen are meaningful for be used for SSSG even if they were not originally designed for that. Finally, we found that our algorithm is very efficient, and it executes a thousand of times faster than the brute force implementation.
\section{Conclusions and Future Work}

In conclusion, we introduce Neural Reward Machines, a neurosimbolic framework providing non-Markovian rewards for non-symbolic-state environments. NRMs possess great versatility and applicability across various learning configurations, with a particular focus here on their capacity for grounding within environments. We address the challenge of semisupervised symbol grounding through temporal specifications and propose an algorithm for identifying Unremovable Reasoning Shortcuts in temporal tasks. We demonstrate the ability of NRMs to integrate the machine's prior knowledge and leverage it to outperform Deep RL methods, achieving nearly equivalent rewards to RMs, even without possessing full prior knowledge. 
We leave exploration of other reasoning-learning scenarios, such as the integration of automata learning with RL, for future research.

\begin{ack}
This work has been partially supported by PNRR MUR project PE0000013-FAIR. 

The work of Francesco Argenziano was carried out when he was
enrolled in the Italian National Doctorate on Artificial Intelligence
run by Sapienza University of Rome.

\end{ack}




\appendix

\section{Training setting}
\paragraph{Implementation and Hardware} The code is entirely written in Python, utilizing PyTorch as our preferred framework, and benefiting from CUDA acceleration. Training was conducted using an NVIDIA 3070 Graphics Card. 

\paragraph{Neural Networks Design} The Critic and Actor networks comprise $3$ fully connected layers with \textit{Tanh} activation functions, each of hidden size equal to $120$. Additionally, the Actor network concludes with an extra \textit{Softmax} activation.
For feature extraction, a CNN was employed, consisting of $2$ convolutional layers, each accompanied by a max pooling layer and \textit{ReLU} activation function. Subsequently, a flatten layer and $2$ cascading fully connected layers follow, concluding with a final \textit{Sigmoid} activation. 
The grounder used in the map environment setting is a Multi Layer Perceptron (MLP) composed of $3$ fully connected layers with dropouts, a \textit{Tanh} between the first $2$ layers and a \textit{Softmax} on the last layer.
In the image environment setting, the grounder takes the form of a CNN, mirroring the structure of the feature extractor. The only distinctions lie in the number of outputs for the second fully connected layer, adjusted to match the number of symbols, and the final activation layer, which employs \textit{Softmax} instead of \textit{Sigmoid}. 
The chosen recurrent network is an LSTM with $two$ hidden layers of $50$ neurons.
\paragraph{Hyperparameters} Each setting was repeated with $5$ different seeds to analyze method variance. Each experiment comprises $10,000$ episodes. Policy updates for the A2C network occur every $5$ steps, while grounder training happen every $120$ episodes and it lasts up to $100$ epochs. The chosen optimizer is Adam, with a learning rate set to $4\mathrm{e}{-4}$. The coefficients used for the Actor's loss, the Critic's loss, and entropy are set to $0.3$, $0.5$, and $1\mathrm{e}{-4}$, respectively.
\paragraph{Other Details} CNNs are exclusively employed in the image environment setting. Images in this setting are reshaped to $64$x$64$ to reduce complexity while preserving information. To enhance plot readability, rewards are averaged using a sliding window of size $100$.

\section{Theorems proof}
Here we demonstrate theorems provided in the main paper.
\paragraph{Theorem 1.} 
Let $D^*_{sym}(L)$ denote the set of all strings over $P$ with maximum length equal to $L$. If $L_1 \leq L_2$, then $\#RS^*(\phi, D^*_{sym}(L_2)) \leq \#RS^*(\phi, D^*_{sym}(L_1))$.  
\begin{proof}
If a mapping \( \alpha \) preserves the target output of the formula on a string of length \( L_2 \), then it also preserves it on the subtrace of length \( L_1 \leq L_2\), the opposite is not generally true, if a mapping preserves the output up to \( L_1 \) it is not said that it preserves it also up to \( L_2\). Therefore the number of reasoning shortcuts can only decrease as we increase the horizon of the complete symbolic support dataset. More formally. Let us denote as $x$ the string of length $L_1$ and $x+y$ the string of length $L_2$. we have
\[\delta^{**}_o(q_0, x+y) = \delta^{**}_o(q_0, (x+y) \circ \alpha) \Rightarrow \delta^{**}_o(q_0, x) = \delta^{**}_o(q_0, x \circ \alpha)\]
Therefore
\[\neg (\delta^{**}_o(q_0, x) = \delta^{**}_o(q_0, x \circ \alpha)) \Rightarrow \neg (\delta^{**}_o(q_0, x+y) = \delta^{**}_o(q_0, (x+y) \circ \alpha))\]
Therefore
\[ \Rightarrow RS(\phi, D^*_{sym}(L_2)) \leq RS(\phi, D^*_{sym}(L_1)) \]
\end{proof}
\paragraph{Corollary 2.}
The number of RS calculated on any complete support with finite horizon $L$ is an upper bound for the number of URS.
$ \#RS^*(\phi)  \leq  \#RS^*(\phi, D^*_{sym}(L)) $ $\forall L < \infty$.
\begin{proof}
Since the number of Unremovable Reasoning Shortcuts is the number of RS on the \textit{complete} (infinite) symbolic dataset, which we denote as $D^*_{sym}(\infty)$, we have
\[\#RS^*(\phi) = \#RS^*(\phi, D^*_{sym}(\infty)) \leq  \#RS^*(\phi, D^*_{sym}(L))\]
because $L < \infty$ and Theorem 1
\end{proof}
\paragraph{Theorem 3.} 
Let $A \subseteq Q$ be the set of absorbing states of $\phi$. \footnote{A state $q \in Q$ is called absorbing if $\delta(q,p) = q$ $\forall p \in P$}. If: (i) a string $x$ has reached an absorbing state, $\delta_t^*(q_0, x) \in A$. (ii) for a certain mapping $\alpha$, also $x \circ \alpha$ has reached an absorbing state, $\delta_t^*(q_0, x \circ \alpha) \in A$. (iii) $\alpha$ is a working map for $x$, namely $\delta_o^{**}(q_0, x) = \delta_o^{**}(q_0, x \circ \alpha)$.
Then $\alpha$ is a working map also for all the strings having $x$ as prefix, namely $\delta^{**}_o(q_0, x + y) = \delta^{**}_o(q_0, (x + y) \circ \alpha) \forall y \in P*$.
\begin{proof}
Since $x$ has reached an absorbing state (point (i)), no matter how we expand the string, the state, and consequently the output that we get in the last step will be always the same. Namely
\[ \delta^*_o(q_0, x) = \delta^*_o(q_0, x+y)) \quad \forall y \in P^*\]
The same holds for the string $x \circ \alpha$, for point (ii).
\[  \delta^*_o(q_0, x \circ \alpha) = \delta^*_o(q_0, (x \circ \alpha) +y))  \forall y \in P^*\]
we can bring +$y$ inside the mapping $\alpha$
\[  \delta^*_o(q_0, x \circ \alpha) = \delta^*_o(q_0, (x+y) \circ \alpha)  \quad \forall y \in P^*\]
for point (iii) we have
\[\delta_o^*(q_0, x) = \delta_o^*(q_0, y\circ \alpha)\]
therefore
\[\delta_o^*(q_0, x+y) = \delta_o^*(q_0, (x+y)\circ \alpha  \quad \forall y \in P^*\]
which demonstrates the thesis
\[ \delta^{**}_o(q_0, x + y) = \delta^{**}_o(q_0, (x + y) \circ \alpha) \quad \forall y \in P^*\]
\end{proof}
\paragraph{Theorem 4.} 
Given that: (i) $\alpha$ is a working map, for the string $x+z$, with $x,z \in P^*$, $\delta_o^{**}(q_0, x+z) = \delta_o^{**}(q_0, (x+z) \circ \alpha)$; (ii) $\exists p \in P$ such that both $x$ and $x+p$ reach the same state and $x \circ \alpha$ and $(x+p) \circ \alpha$ reach the same state, $\delta_t^*(q_0, x) = \delta_t^*(q_0, x+p) = q \quad \wedge \quad \delta_t^*(q_0, x) = \delta_t^*(q_0, x+p)$.
Then $\alpha$ works for the string $x+y+z$, $\delta_o^{**}(q_0, x+y+z) = \delta_o^{**}(q_0, (x+y+z) \circ \alpha)$, $\forall y \in p^*$ \footnote{with $p^*$ being the set of strings obtained repeating the symbol $p$ an arbitrary number of times.}.
\begin{proof}
We can write point (i) separating the result for string $x$ and $z$ in the following way
\begin{equation}
\begin{array}{l}
    \delta_o^{**}(q_0, x) = \delta_o^{**}(q_0, x \circ \alpha)\quad \wedge \\
    
    \wedge \quad \delta_o^{**}(\delta_t^*(q_0,x), z) = \delta_o^{**}(\delta_t^*(q_0, x \circ \alpha), z \circ \alpha)
\end{array}
\end{equation}
Analogously, we can demonstrate the thesis by proving
\begin{equation}
\begin{array}{l}
    \delta_o^{**}(q_0, x+y) = \delta_o^{**}(q_0, (x+y) \circ \alpha) \quad \wedge \\
    \wedge \quad \delta_o^{**}(\delta_t^*(q_0, x+y), z) = \delta_o^{**}(\delta_t^*(q_0, (x+y)\circ \alpha), z \circ \alpha)
\end{array}
\end{equation}
We start by proving the first conjunct of (10).

Because of (9) (first conjunct) we have
\[\delta_o^*(q_0, x) = \delta_o^*(q_0, x \circ \alpha)\]
For point (ii) we have 
\[\delta_o^*(q_0, x) = \delta_o^*(q_0, x+y) \quad \forall y \in p^*\]
and also
\[\delta_o^*(q_0, x \circ \alpha) = \delta_o^*(q_0, (x+y) \circ \alpha) \quad \forall y \in p^*\]
combining the last three statement we have 
\[\delta_o^*(q_0, x+y) = \delta_o^*(q_0, (x+y) \circ \alpha) \quad \forall y \in p^*\]
which demonstrate the first conjunct of (10).

We proceed demonstrating the second part of (10).
For point (ii) we have
\[\delta_t^*(q_0, x+y) = \delta_t^*(q_0, x)\]
and also that
\[\delta_t^*(q_0, (x+y) \circ \alpha) = \delta_t^*(q_0, x \circ \alpha)\]
replacing in the second conjunct of (10) we eliminate the y and we have the second conjunct of (9), which is true by hypothesis.

Hence the thesis is proven.
\end{proof}

\begin{table*}[t!]
    \centering
    \begin{tabular}{lcp{0.2\linewidth}p{0.2\linewidth}p{0.1\linewidth}cc}
        \toprule
        Task & Class & Pattern & Formula & \#RS* & Alg. 1 (s) & Brute-force Alg. (s)\\
        \midrule
        1 & 1 & Visit($a, b$) & F(a) $\wedge$ F(b) & $54$ & $0.198$ & $270.364$\\ \\
        2 & 1 & Visit($a, b, c$) & F(a) $\wedge$ F(b) $\wedge$ F(c) & $24$ & $1.020$ & $668.362$\\ \\
        3 & 1 & Sequenced\_Visit($a, b$) & F(a $\wedge$ F(b)) & $27$ & $0.181$ & $214.006$\\ \\
        4 & 1 & Sequenced\_Visit($a, b$) + Visit($c$) & F(a $\wedge$ F(b)) $\wedge$ F(c) & $4$ & $0.877$ & $585.556$\\ \\
        5 & 2 & Visit($a, b$) + Global\_Avoidance($c$) & F(a) $\wedge$ F(b) $\wedge$ G($\neg$c) & $8$ & $0.204$ & $416.242$\\ \\
        6 & 2 & Visit($a, b$) + Global\_Avoidance($c, d$) & F(a) $\wedge$ F(b) $\wedge$ G($\neg$c) $\wedge$ G($\neg$d) & $8$ & $0.181$ & $431.972$\\ \\
        7 & 2 & Sequenced\_Visit($a, b$) + Global\_Avoidance($c$) & F(a $\wedge$ F(b)) $\wedge$ G($\neg$c) & $4$ & $0.194$ & $369.701$\\ \\
        8 & 2 & Sequenced\_Visit($a, b$) + Global\_Avoidance($c, d$) & F(a $\wedge$ F(b)) $\wedge$ G ($\neg$c) $\wedge$ G($\neg$d) & $4$ & $0.187$ & $381.943$ \\ \\
        \bottomrule
    \end{tabular}
    \caption{Complete analysis of formulas selected as RL tasks}
    \label{tab:groundabilityfull}
\end{table*}

\section{Experiments}
\begin{figure*}
        \centering
\subfigure[]{
    \includegraphics[width=0.35\textwidth]{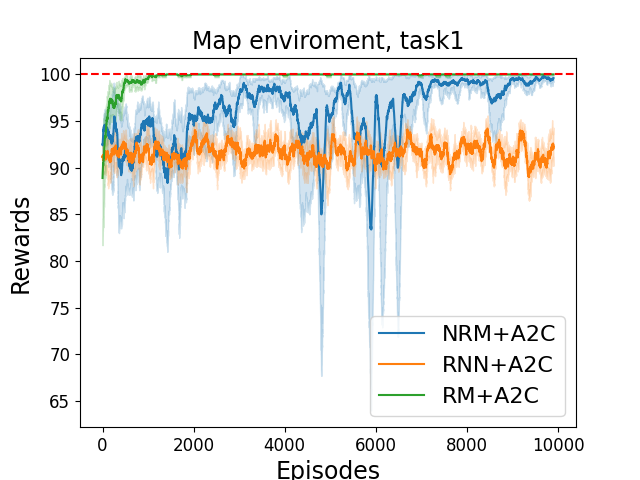}
    }
\subfigure[]{
    \includegraphics[width=0.35\textwidth]{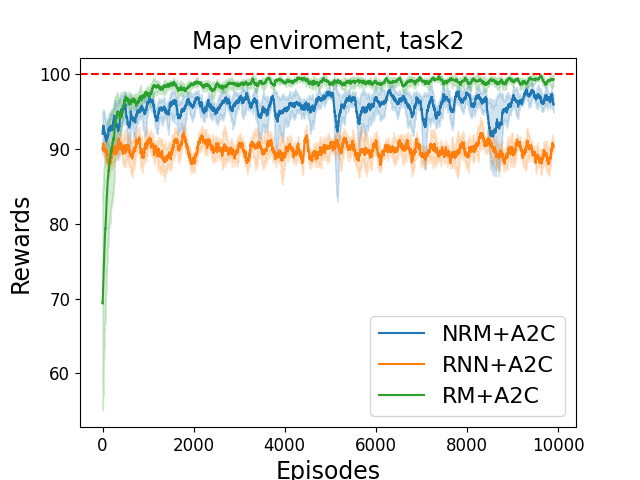}
    }
\subfigure[]{
    \includegraphics[width=0.35\textwidth]{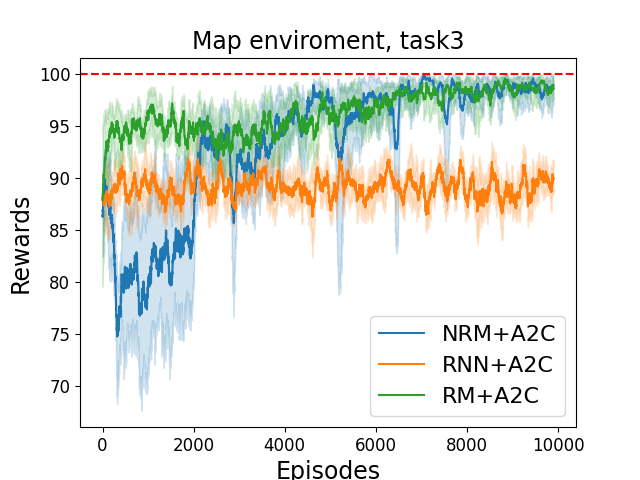}
}
\subfigure[]{
    \includegraphics[width=0.35\textwidth]{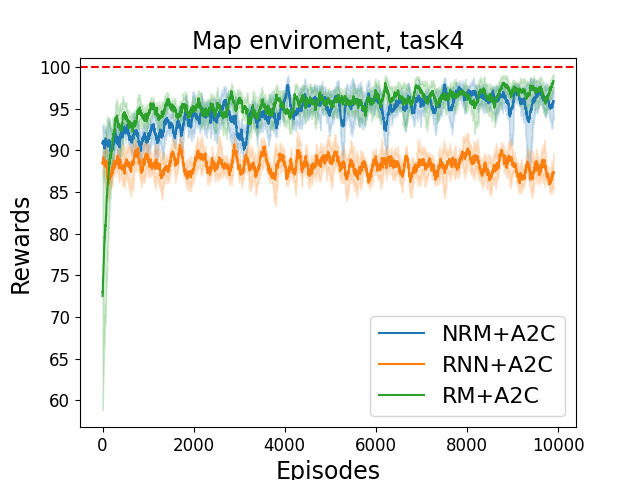}
}
\subfigure[]{
    \includegraphics[width=0.35\textwidth]{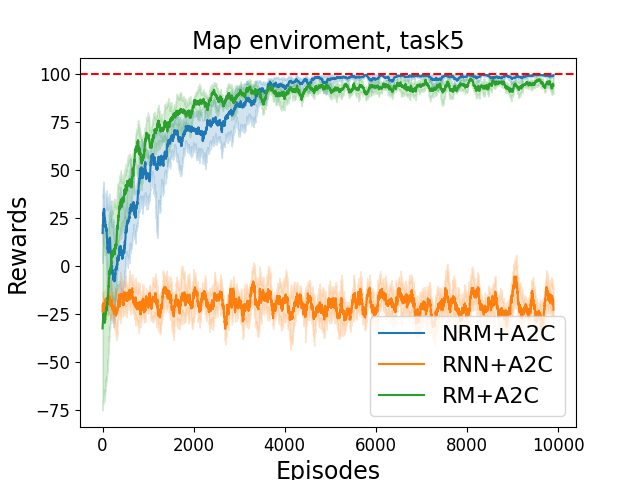}
    }
\subfigure[]{
    \includegraphics[width=0.35\textwidth]{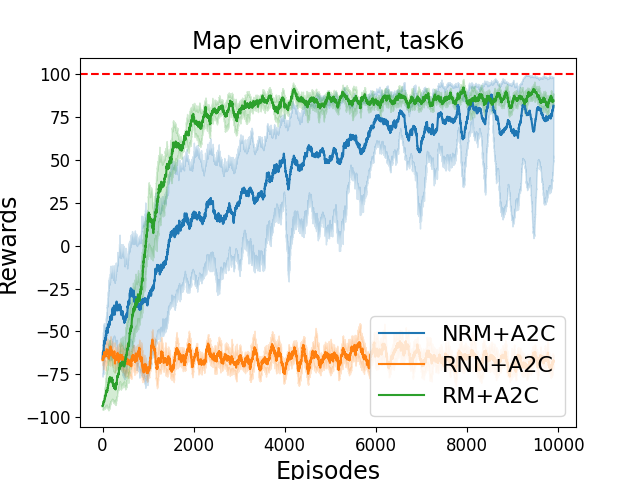}
    }
\subfigure[]{
    \includegraphics[width=0.35\textwidth]{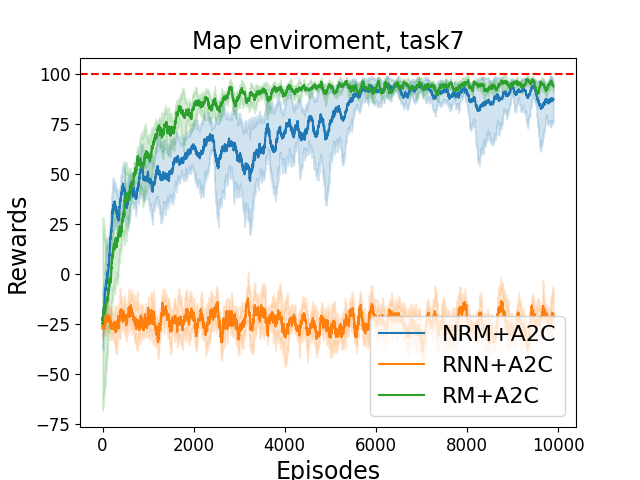}
}
\subfigure[]{
    \includegraphics[width=0.35\textwidth]{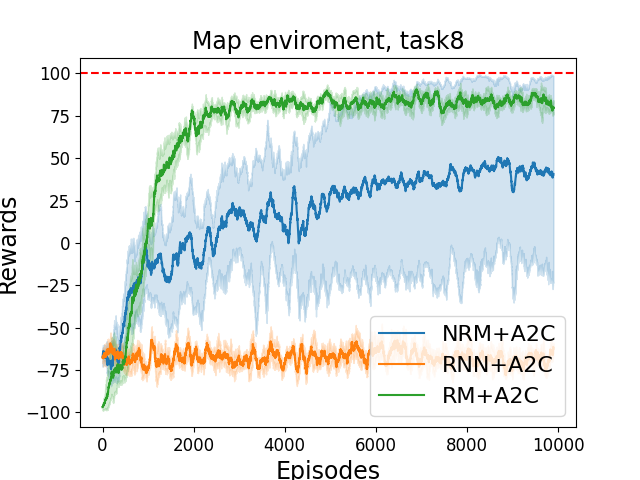}
}
\caption{Training rewards for tasks of Class 1 and 2 on map environment.}
\label{fig:maprewards}
\end{figure*}

\begin{figure*}
        \centering
\subfigure[]{
    \includegraphics[width=0.35\textwidth]{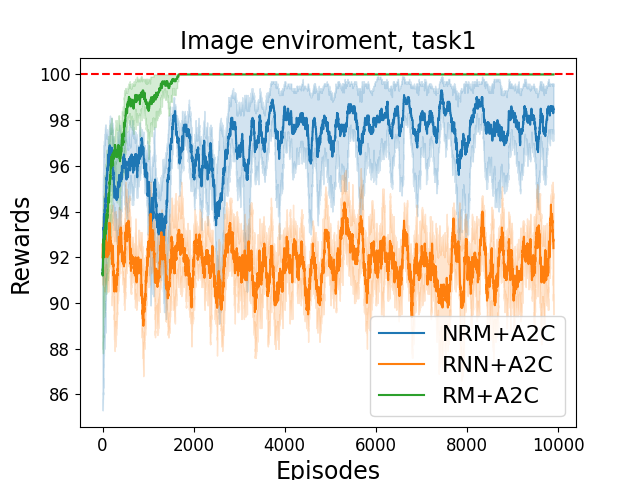}
    }
\subfigure[]{
    \includegraphics[width=0.35\textwidth]{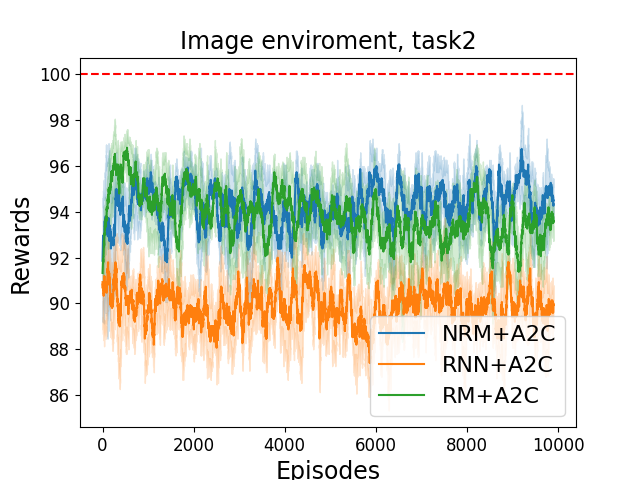}
    }
\subfigure[]{
    \includegraphics[width=0.35\textwidth]{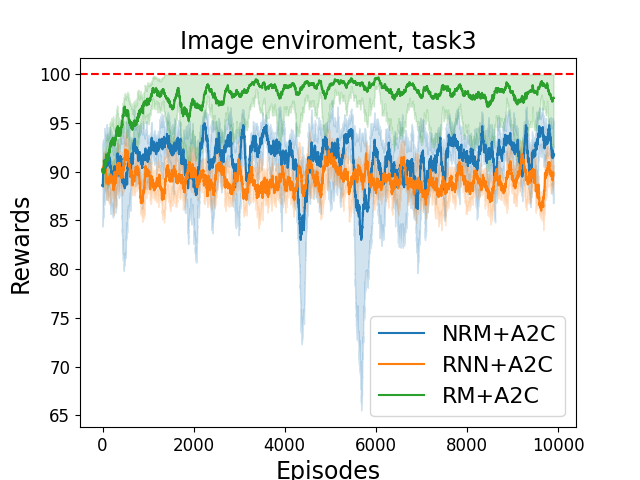}
}
\subfigure[]{
    \includegraphics[width=0.35\textwidth]{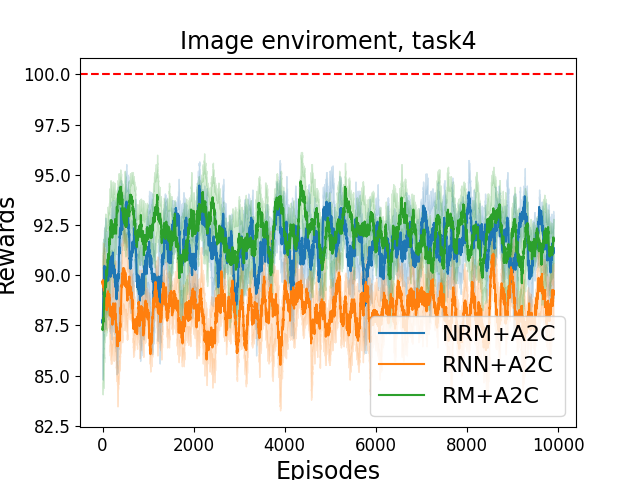}
}
\subfigure[]{
    \includegraphics[width=0.35\textwidth]{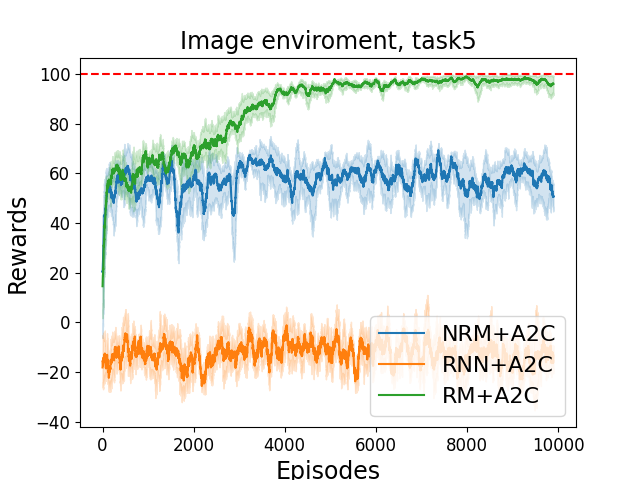}
    }
\subfigure[]{
    \includegraphics[width=0.35\textwidth]{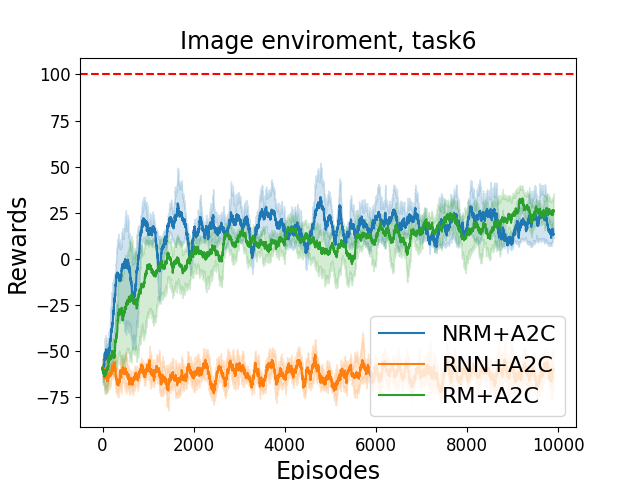}
    }
\subfigure[]{
    \includegraphics[width=0.35\textwidth]{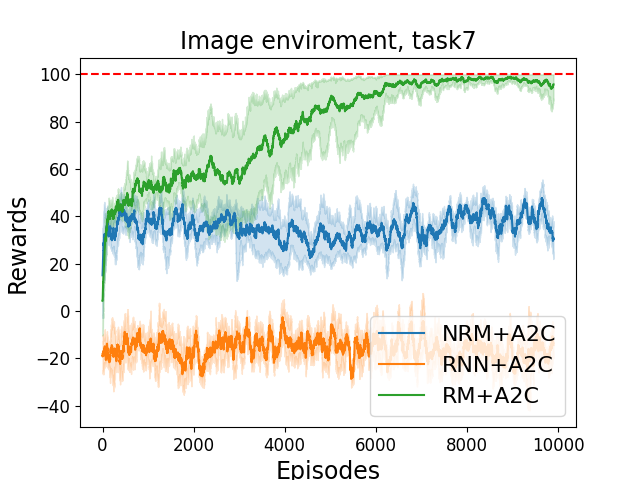}
}
\subfigure[]{
    \includegraphics[width=0.35\textwidth]{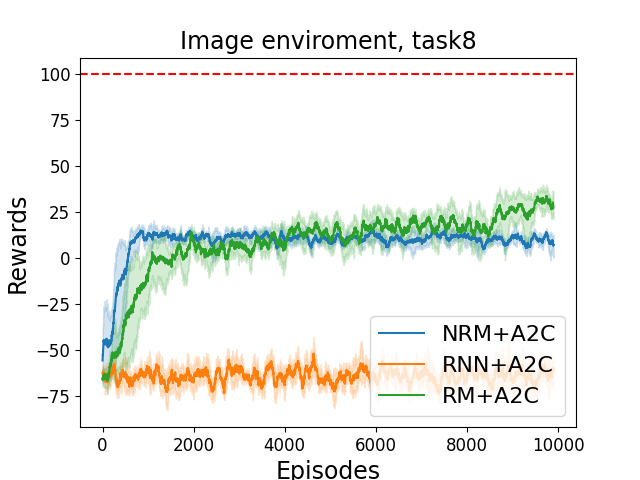}
}
\caption{Training rewards for tasks of Class 1 and 2 on image environment.}
\label{fig:imagerewards}
\end{figure*}

Finally we report the results obtained in the single tasks considered, which are reported aggregated by task class in the main paper for space reasons.
In Table \ref{tab:groundabilityfull} we show the groundability analysis on all the formulas. We report the  number of URS found ($\#RS^*$) and the comparison in execution time between our algoritm (1) and a brute-force one. We show in Figure \ref{fig:maprewards} and  \ref{fig:imagerewards} the training rewards obtained in the single tasks by the three method tested (RNN, NRM and RM) in the map and the image environment respectively.

\newpage

\newpage

\end{document}